\documentclass[letterpaper]{article}
\usepackage[preprint]{aaai2027}
\usepackage[hyphens]{url}  
\usepackage{graphicx} 
\usepackage{natbib}  
\usepackage{caption} 
\usepackage{algorithm}
\usepackage{algorithmic}

\usepackage{amsmath} 
\usepackage{amssymb}
\usepackage[most]{tcolorbox}
\usepackage[table]{xcolor}       
\usepackage{colortbl}            
\usepackage{multirow}
\usepackage{pifont}

\newtcolorbox{promptbox}[1]{%
  enhanced,
  breakable,
  colback=gray!3,
  colframe=gray!45,
  boxrule=0.5pt,
  arc=2pt,
  left=6pt, right=6pt, top=5pt, bottom=5pt,
  title=\textbf{#1},
  fonttitle=\normalsize,
  before skip=4pt,
  after skip=6pt
}
\usepackage{braket}              
\usepackage{newfloat}
\usepackage{listings}
\DeclareCaptionStyle{ruled}{labelfont=normalfont,labelsep=colon,strut=off} 
\floatstyle{ruled}
\newfloat{listing}{tb}{lst}{}
\floatname{listing}{Listing}

\usepackage{booktabs}

\title{Shared Prefixes, Better Credit: Adaptive Routing for Multi-Agent Reasoning}
\author{
    Yiqing Liu,
    Zihao Wang,
    Hantao Yao,
    Wu Liu,
    Yongdong Zhang
}

\affiliations{
    University of Science and Technology of China
}

\begin{document}
\maketitle

\begin{abstract}

Multi-agent reasoning (MAR) improves reasoning reliability through iterative solution exchange and refinement. Existing adaptive MAR methods typically learn routing decisions from query-level labels or trajectory-level returns, but such coarse supervision cannot accurately estimate the state-conditioned utility of individual operators in multi-step collaboration. 
We propose \textbf{TreeCredit}, a shared-prefix credit assignment framework for efficient adaptive MAR. Its core insight is to estimate operator utility through \emph{state-matched downstream comparisons}, rather than directly attributing trajectory-level outcomes to preceding decisions. TreeCredit constructs shared-prefix collaboration trees by expanding candidate operators from the same intermediate state and assigns each state--operator pair a correctness-prioritized suffix credit based on the terminal correctness and cumulative additional cost of its complete continuation. These structured credits are converted into state-local operator preferences to train a lightweight pairwise state router, which dynamically selects the next admissible operator during inference. 
Experiments on six reasoning benchmarks show that TreeCredit modestly improves accuracy while substantially reducing inference cost, achieving a better accuracy--cost trade-off than representative MAR methods.


\end{abstract}

\section{Introduction}

Large language model based multi-agent reasoning (MAR) enables multiple agents to exchange complementary analyses, refine intermediate solutions, and aggregate diverse opinions for complex reasoning tasks~\cite{qian2024chatdev,hong2024metagpt,chen2024agentverse}.
By combining diverse reasoning capabilities with structured interaction, MAR systems improve robustness and solution quality over isolated inference.
However, these gains often require additional model calls, message exchanges, and agent coordination, resulting in substantial inference overhead that increases with the scale and structural complexity of collaboration.
Improving reasoning quality while controlling collaboration cost has therefore become a central objective for efficient multi-agent reasoning.

\begin{figure}
	\centering
    \includegraphics[
    width=0.9\columnwidth,
    height=0.28\textheight]{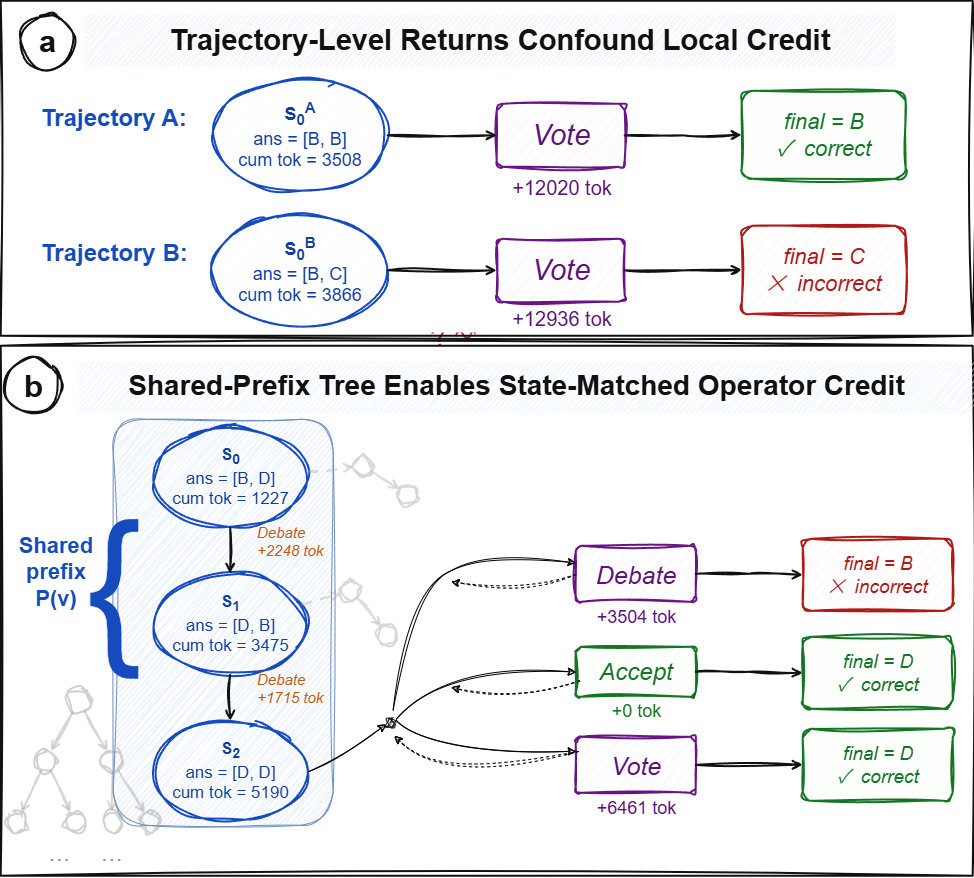}
    \caption{Motivation for state-matched operator credit.
(a) The same operator yields different outcomes across states.
(b) Different operators at the same state produce different correctness and costs.}
 \label{fig:1}
\end{figure}

To address these limitations, recent studies have shifted from predefined collaboration workflows toward adaptive routing and orchestration. MasRouter~\cite{yue2025masrouter} and DAAO~\cite{su2026daao} dynamically configure collaboration modes, agent roles, model assignments, and workflow depth based on query characteristics or estimated difficulty. MaAS~\cite{zhang2025maas} and BAMAS~\cite{yang2026bamas} further construct query-dependent agentic architectures under performance and budget constraints. Graph-based approaches adapt communication structures by learning interaction topologies, sparsifying edges, or controlling agent participation and information visibility~\cite{zhuge2024gptswarm,zhang2025gdesigner,yang2026agenttailor,wang2026rumad}. More recently, sequential routing methods revise collaboration through state-conditioned policies that determine whether to continue debate, invoke additional models, or modify communication patterns~\cite{fan2026imad,zhang2025routerr1,lu2026dytopo}. Collectively, these methods demonstrate the potential of adaptive orchestration to improve the trade-off between reasoning accuracy and inference cost. 
However, intermediate routing decisions are often supervised using query-level labels or trajectory-level returns, which provide only indirect and coarse credit. Since the outcome is jointly determined by a sequence of interdependent decisions, successful trajectories may still contain unnecessary or harmful choices, whereas failed trajectories may include locally beneficial ones.
Therefore, how to estimate the state-conditioned utility of intermediate operators is a critical aspect for learning efficient and reliable multi-agent routing policies.

Our key insight is to learn operator utility through \emph{state-matched downstream comparisons}, rather than directly attributing trajectory-level outcomes to preceding routing decisions. 
As illustrated in Figure~\ref{fig:1}, applying different operators to the same intermediate state can lead to substantially different downstream correctness and inference costs, indicating that the preferred operator depends on the reached collaboration state. 
We therefore evaluate candidate operators from the same shared-prefix state, holding the query and interaction history. 
Each operator is assessed by the final correctness and additional inference cost of the complete continuation it induces. 
The shared prefix controls variation caused by preceding execution histories, while the complete suffix preserves the delayed accuracy--cost consequences of the current choice, as shown in Figure~\ref{fig:1}(b). 
These state-matched comparisons transform coarse trajectory-level feedback into fine-grained local preferences for learning reliable state-aware routing policies.

Based on this insight, we propose \textbf{TreeCredit}, a shared-prefix credit assignment framework for adaptive multi-agent reasoning. 
Through \emph{Shared-Prefix Tree Expansion}, TreeCredit constructs independently sampled offline collaboration trees by expanding all admissible operators as sibling branches from the same intermediate state. 
Because these branches share an identical query, interaction history, and computational prefix, their downstream outcomes provide state-matched evidence for comparing operator utility, with each sampled state realization contributing independent supervision. 
\emph{Correctness-Prioritized Suffix Credit} then assigns each state--operator pair a structured credit based on the terminal correctness and cumulative additional cost of the best complete suffix represented in the sampled tree. 
\emph{State-Local Preference Extraction} converts the resulting credits into pairwise operator preferences, prioritizing correct continuations and using lower cost to distinguish alternatives with equal correctness. 
Finally, \emph{Pairwise State Routing} trains a lightweight router from these preferences to select the highest-scoring admissible operator at each state. 
The collaboration trees are used only for offline supervision construction, whereas inference follows a single adaptive trajectory until an acceptance or voting operator terminates reasoning.

Our major contributions can be summarized as follows:
\begin{itemize}
\item We identify state-conditioned credit assignment as a key challenge in adaptive MAR and introduce \emph{state-matched downstream comparison} to estimate operator utility from comparable complete continuations.
\item We propose \textbf{TreeCredit}, a shared-prefix credit assignment framework that constructs state-local preference targets by comparing the best complete continuations observed for sibling operators under a shared-prefix state.

\item Experiments across multiple complex reasoning benchmarks demonstrate that TreeCredit effectively reduces inference cost while improving task accuracy.
\end{itemize}

\section{{Related Work}}
\paragraph{Efficient and Adaptive Multi-Agent Collaboration.}
Existing methods improve multi-agent efficiency by adapting interaction depth, model allocation, agent participation, or communication structure.
Free-MAD reduces redundant consensus seeking, while iMAD determines whether additional debate is warranted~\cite{cui2026free,fan2026imad}.
At the query and workflow level, MasRouter, DAAO, MaAS, and MAS-Orchestra construct query-dependent model combinations, agentic architectures, or collaboration workflows, while BAMAS additionally considers computation budgets~\cite{yue2025masrouter,su2026daao,zhang2025maas,ke2026masorchestra,yang2026bamas}.
Structure-oriented methods optimize communication graphs, edge activation, agent roles, or model mixtures~\cite{zhang2025gdesigner,zhang2026radar,yang2026agenttailor,li2025omac,yao2026hieramas,li2026maca}.
During execution, DyTopo, Router-R1, Optimal-Agent-Selection (STRMAC), OI-MAS, and RUMAD adapt communication topology, agent or model selection, and debate behavior according to the evolving interaction state~\cite{lu2026dytopo,zhang2025routerr1,wang2025strmac,wang2026oimas,wang2026rumad}.
TreeCredit builds on this direction by treating heterogeneous collaboration mechanisms as callable operators and learning which operator to apply at each reached state.
\paragraph{Credit Assignment for Multi-Agent Orchestration.}
Learning an adaptive orchestration policy requires attributing downstream performance to individual routing decisions.
Existing methods commonly optimize scalar rewards defined over steps or complete trajectories.
BAMAS combines task performance with budget efficiency~\cite{yang2026bamas}; Router-R1 combines format, final-answer, and model-usage signals~\cite{zhang2025routerr1}; and RUMAD uses step- and episode-level objectives involving accuracy, consensus, efficiency, and sparsity~\cite{wang2026rumad}.
Although effective for end-to-end optimization, these objectives do not directly rank the alternative operators available at the same state because a trajectory outcome reflects the combined effects of multiple interdependent decisions.
LEMON addresses a related problem by counterfactually editing role, capacity, or dependency fields in an executable orchestration specification and assigning reward contrasts to the edited spans~\cite{chen2026lemon}.
TreeCredit instead compares \emph{runtime operator choices} expanded from an identical interaction history, evaluates their downstream suffixes using correctness and additional inference cost, and converts these state-matched comparisons into local pairwise preferences.


\begin{figure*}
	\centering
	\includegraphics[width=0.9\linewidth]{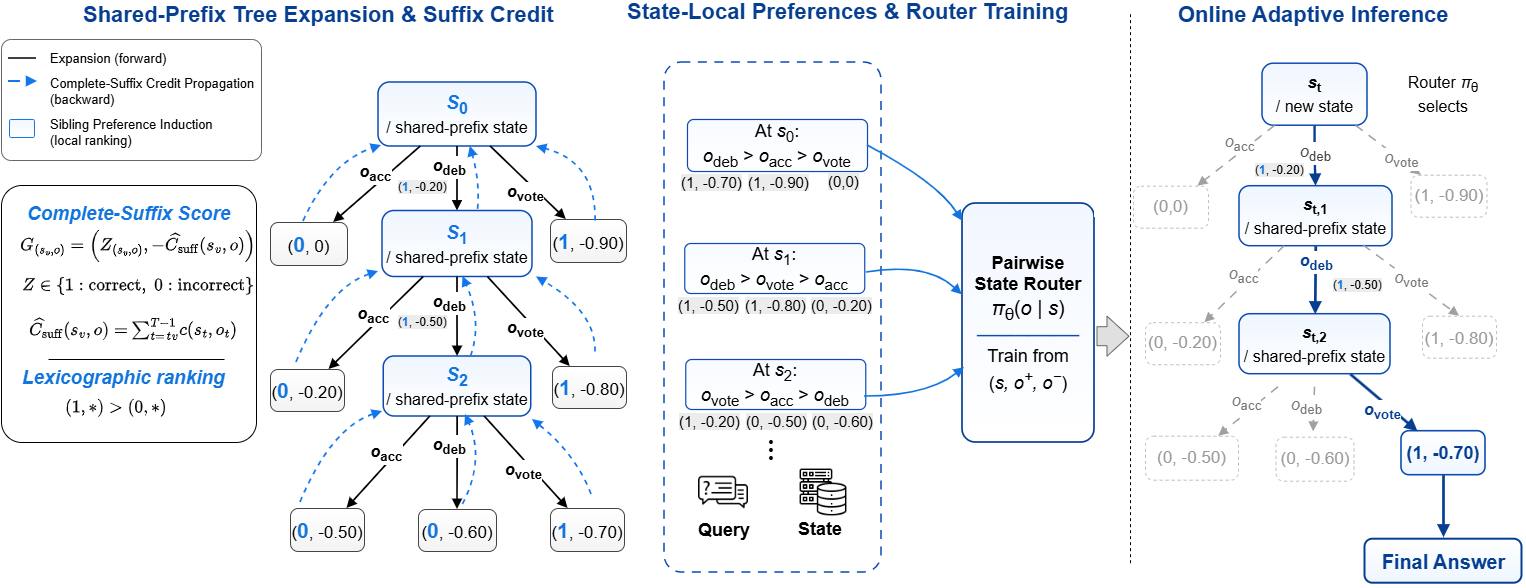}
    \caption{Overview of TreeCredit. Shared-prefix collaboration trees
are expanded offline to assign correctness-prioritized suffix credits
and extract state-local operator preferences. These preferences train
a pairwise state router that follows a single adaptive trajectory
during inference.}
 \label{fig:2}
\end{figure*}

\section{Formalization}
\label{sec:formalization}

\paragraph{Dynamic Multi-Agent Reasoning.}
Given a query $q$, multi-agent reasoning evolves over a sequence of collaboration
states: $s_t=\left(q,H_t,B_t\right)$,
where $H_t$ contains the observable interaction history, including candidate
answers, agent responses, and previously executed operators, and $B_t$ denotes
the remaining inference budget.

Defining $\mathbb{O}=\left\{o_1,\ldots,o_L\right\}$ be a finite catalog of heterogeneous collaboration operators. 
Each operator
specifies a participating agent set, an interaction protocol, an induced state
transition, and an inference cost. 
At state $s_t$, the router selects an admissible operator according to Eq.~\eqref{eq:routing_policy},
\begin{equation}
o_t
\sim
\pi_{\theta}
\left(
\cdot
\mid
s_t
\right),
\qquad
o_t
\in
\mathbb{O}(s_t)
\subseteq
\mathbb{O},
\label{eq:routing_policy}
\end{equation}
where $\mathbb{O}(s_t)$ is determined by the collaboration protocol. 
Executing $o_t$ induces a following stochastic transition:
\begin{equation}
s_{t+1}
\sim
P
\left(
\cdot
\mid
s_t,
o_t
\right),
\qquad
B_{t+1}
=
B_t-c(s_t,o_t),
\label{eq:state_transition}
\end{equation}
and incurs an immediate cost $c(s_t,o_t)\geq 0$.

A complete collaboration trajectory is defined as $\tau$,
\begin{equation}
\tau
=
\left(
s_0,
o_0,
s_1,
\ldots,
o_{T-1},
s_T
\right),
\label{eq:trajectory}
\end{equation}
where $s_T$ is terminal. 
The final answer and total inference cost are:
\begin{equation}
\hat{y}_{\tau}
=
f(s_T),
\qquad
C(\tau)
=
\sum_{t=0}^{T-1}
c(s_t,o_t),
\label{eq:trajectory_outcome}
\end{equation}
where $f$ extracts the final prediction from the terminal state.

In this work, we instantiate the finite catalog of heterogeneous collaboration operators as,
\begin{equation}
\mathbb{O}
=
\left\{
o_{\mathrm{acc}},
o_{\mathrm{deb}},
o_{\mathrm{vote}}
\right\},
\label{eq:operator_instantiation}
\end{equation}
where $o_{\mathrm{acc}}$ accepts the current answer and terminates reasoning,
$o_{\mathrm{deb}}$ performs another heterogeneous debate round and produces a
successor state, and $o_{\mathrm{vote}}$ aggregates the current candidate
answers and terminates.

\paragraph{System-Level Design Objective.}
Defining $\mathcal{D}=\left\{(q_i,y_i)\right\}_{i=1}^{M}$ as a labeled benchmark, and $U(\hat{y},y)=\mathbb{I}[\hat{y}=y]\in\{0,1\}$ as the correctness indicator. 
We formulate adaptive MAR as minimizing expected inference cost while preserving accuracy. Rather than solving this constrained objective directly, 
TreeCredit uses correctness-prioritized state-local preferences as a practical surrogate, ranking correct continuations before comparing costs.
\begin{equation}
\begin{aligned}
\min_{\theta}\quad
&
\mathbb{E}_{(q,y)\sim\mathcal{D},\,
\tau\sim\pi_{\theta}}
\left[
C(\tau)
\right]
\\
\mathrm{s.t.}\quad
&
\mathbb{E}_{(q,y)\sim\mathcal{D},\,
\tau\sim\pi_{\theta}}
\left[
U(\hat{y}_{\tau},y)
\right]
\geq
A_{\mathrm{ref}}-\epsilon,
\end{aligned}
\label{eq:routing_objective}
\end{equation}
where $A_{\mathrm{ref}}$ is a reference accuracy and $\epsilon\geq 0$ is the
allowed accuracy tolerance. 
Note that this equation specifies the system-level design objective, while the preference-learning objective introduced later serves as its state-local surrogate.

A terminal outcome evaluates the complete trajectory but does not identify the downstream utility of any individual state--operator decision. 
TreeCredit addresses this credit-assignment gap by comparing candidate operators from shared-prefix states and deriving state-local preferences from their complete downstream outcomes.

\section{Method}
\label{sec:method}
As shown in Figure~\ref{fig:2}, TreeCredit learns the state-conditioned routing policy $\pi_\theta$ from offline collaboration trees through four modules: \emph{Shared-Prefix Tree Expansion}, \emph{Correctness-Prioritized Suffix Credit}, \emph{State-Local Preference Extraction}, and \emph{Pairwise State Routing}.  
The trees are used only to construct supervision; at inference time, TreeCredit follows a single adaptive collaboration trajectory.

\paragraph{Shared-Prefix Tree Expansion:}\label{sec:shared_prefix_trees}
For each labeled example $(q_i,y_i)\in\mathcal{D}$, TreeCredit constructs $K$ independently sampled collaboration trees:
\begin{equation}
\mathcal{T}_{i}^{(k)}
=
\left(
\mathcal{V}_{i}^{(k)},
\mathcal{E}_{i}^{(k)}
\right),
\qquad
k\in\{1,\ldots,K\}.
\label{eq:tree_definition}
\end{equation}
Each node $v\in\mathcal{V}_{i}^{(k)}$ represents a reached collaboration state $s_v=(q_i,H_v,B_v)$. Independent trees use different settings decoding randomness and therefore provide diverse state realizations for offline credit construction. 
Each sampled state realization is treated independently when generating local preference examples.

At every non-terminal node $v$, TreeCredit expands all admissible operators
from the same state:
\begin{equation}
(s_v,o)
\;\xrightarrow{\;\mathrm{Exec}\;}
\begin{cases}
\text{terminal state }s_{\ell},
&
o\in\{o_{\mathrm{acc}},o_{\mathrm{vote}}\},
\\[2mm]
\text{child state }s_u,
&
o=o_{\mathrm{deb}}.
\end{cases}
\label{eq:tree_expansion}
\end{equation}
Each edge is labeled by the executed operator and its immediate cost
$c(s_v,o)$. Debate expansion continues recursively until a terminal operator is
executed or the maximum depth $D_{\max}$ is reached.

A branch initiated by operator $o$ at node $v$ can be decomposed as:
\begin{equation}
\tau(v,o)
=
P(v)
\oplus
(s_v,o)
\oplus
S(v,o),
\label{eq:trajectory_decomposition}
\end{equation}
where $P(v)$ is the common prefix from the root to $v$, and $S(v,o)$ is the
remaining downstream suffix after executing $o$.
For any sibling operators
$o_a,o_b\in\mathbb{O}(s_v)$, the accumulated prefix cost is identical.
Therefore, the difference between those two operators is:
\begin{equation}
\begin{aligned}
&
\left[
C_{\mathrm{pre}}(v)
+
C_{\mathrm{suf}}(s_v,o_a)
\right]
-
\left[
C_{\mathrm{pre}}(v)
+
C_{\mathrm{suf}}(s_v,o_b)
\right]
\\
&\qquad
=
C_{\mathrm{suf}}(s_v,o_a)
-
C_{\mathrm{suf}}(s_v,o_b),
\end{aligned}
\label{eq:prefix_cancellation}
\end{equation}
where $C_{\mathrm{pre}}(\cdot)$ and $C_{\mathrm{suf}}(\cdot)$ denote the cost of prefix and suffix operators, respectively.
The shared prefix controls differences in preceding execution histories, while suffix outcomes provide state-matched estimates of the downstream utility of candidate operators. 
This comparison estimates the utility of selecting an operator together with its complete continuation, rather than its isolated one-step causal effect.

\paragraph{Correctness-Prioritized Suffix Credit}
\label{sec:suffix_credit}
For each expanded state--operator pair $(s_v,o)$, TreeCredit defines a
structured suffix credit $G(s_v,o)$,
\begin{equation}
G(s_v,o)
=
\left(
Z(s_v,o),
-
C_{\mathrm{suf}}(s_v,o)
\right),
\label{eq:suffix_credit}
\end{equation}
where $Z(s_v,o)\in\{0,1\}$ is the correctness of the terminal prediction
reached after selecting $o$, and $C_{\mathrm{suf}}(s_v,o)$ is the total cost from $s_v$ through $o$ to termination. Both dimensions are written so that a larger value is preferred.

Credits are compared lexicographically:
\begin{equation}
(z,-c)
\succ_{\mathrm{lex}}
(z',-c')
\iff
\left[
z>z'
\right]
\vee
\left[
z=z'
\land
c<c'
\right].
\label{eq:credit_ordering}
\end{equation}
This correctness-prioritized ordering prevents a cheaper but incorrect branch from dominating a correct branch, while favoring lower inference cost among continuations with equal correctness.

For a terminal operator, the suffix credit is observed directly:
\begin{equation}
G(s_v,o)
=
\left(
U(\hat{y}_{s_v,o},y_i),
-
c(s_v,o)
\right),
o\in
\left\{
o_{\mathrm{acc}},
o_{\mathrm{vote}}
\right\}.
\label{eq:terminal_credit}
\end{equation}

If $o_{\mathrm{deb}}$ transitions to successor state $s_u$, its credit is
backed up from the best complete suffix reachable within the sampled tree:
\begin{equation}
G(s_v,o_{\mathrm{deb}})
=
\max_{o'\in\mathbb{O}(s_u)}^{\mathrm{lex}}
\left(
Z(s_u,o'),
-
\left[
c(s_v,o_{\mathrm{deb}})
+
C_{\mathrm{suf}}(s_u,o')
\right]
\right).
\label{eq:recursive_backup}
\end{equation}
This optimal-continuation backup is used only for offline supervision
construction and estimates the potential downstream utility of selecting
debate at $s_v$ under the best continuation represented in the sampled tree.

\paragraph{State-Local Preference Extraction}
\label{sec:preference_extraction}
For any two admissible operators with strictly ordered suffix credits,
TreeCredit constructs a state-local preference:
\begin{equation}
o_a
\succ_{s_v}
o_b
\iff
G(s_v,o_a)
\succ_{\mathrm{lex}}
G(s_v,o_b),
o_a,o_b\in\mathbb{O}(s_v).
\label{eq:local_preference}
\end{equation}
Using all strictly ordered pairs preserves the full state-local ranking rather
than comparing only the best operator against the rest. Aggregating these
comparisons over all independently sampled states, trees, and examples yields
\begin{equation}
\mathcal{D}_{\mathrm{pref}}
=
\left\{
(s_j,o_j^{+},o_j^{-},w_j)
\right\}_{j=1}^{J},
\label{eq:preference_dataset}
\end{equation}
where $o_j^{+}\succ_{s_j}o_j^{-}$, $J=|\mathcal{D}_{\mathrm{pref}}|$, and
$w_j>0$ is a weight that emphasizes correctness-changing comparisons and
balances opposite preference directions within each unordered operator pair.

\paragraph{Pairwise State Routing}
\label{sec:router_learning}
The router operates on a compact state representation
\begin{equation}
\mathbf{x}_v
=
\phi(s_v)
\in
\mathbb{R}^{d},
\label{eq:state_features}
\end{equation}
where $\phi(\cdot)$ extracts process-level features from the complete observable
state $s_v$, including the query, answer agreement, answer transitions,
interaction stability, debate progress, accumulated computation, and remaining
budget.

A lightweight multilayer perceptron maps the state representation to operator
logits:
\begin{equation}
\mathbf{u}_{\theta}(s_v)
=
\operatorname{MLP}_{\theta}
\left(
\mathbf{x}_v
\right)
\in
\mathbb{R}^{|\mathbb{O}|}.
\label{eq:router_logits}
\end{equation}
Let $u_{\theta}(s_v,o)$ denote the logit assigned to operator $o$. Masking
inadmissible operators gives
\begin{equation}
\pi_{\theta}(o\mid s_v)
=
\frac{
\exp\left(
u_{\theta}(s_v,o)
\right)
}{
\displaystyle
\sum_{\tilde{o}\in\mathbb{O}(s_v)}
\exp\left(
u_{\theta}(s_v,\tilde{o})
\right)
},
\qquad
o\in\mathbb{O}(s_v).
\label{eq:masked_policy}
\end{equation}

For the $j$-th preference example
$(s_j,o_j^{+},o_j^{-},w_j)\in\mathcal{D}_{\mathrm{pref}}$, define the logit
margin
\begin{equation}
\Delta_{\theta,j}
=
u_{\theta}(s_j,o_j^{+})
-
u_{\theta}(s_j,o_j^{-}).
\label{eq:preference_margin}
\end{equation}
The router is trained using the weighted Bradley--Terry loss
\begin{equation}
\mathcal{L}_{\mathrm{pref}}(\theta)
=
\frac{
\displaystyle
\sum_{j=1}^{J}
w_j
\log
\left(
1+\exp(-\Delta_{\theta,j})
\right)
}{
\displaystyle
\sum_{j=1}^{J}w_j
}.
\label{eq:preference_loss}
\end{equation}
This objective learns state-local operator preferences without requiring a
globally comparable scalar value function across different collaboration
states.

At inference step $t$, TreeCredit directly selects the highest-scoring
admissible operator:
\begin{equation}
o_t
=
\operatorname*{arg\,max}_{o\in\mathbb{O}(s_t)}
u_{\theta}(s_t,o).
\label{eq:inference_routing}
\end{equation}
Selecting $o_{\mathrm{deb}}$ produces a successor state, whereas selecting
$o_{\mathrm{acc}}$ or $o_{\mathrm{vote}}$ terminates the trajectory and returns the final prediction $\hat{y}_{\tau}=f(s_T)$.

\begin{table*}[t]
\centering
\footnotesize
\setlength{\tabcolsep}{2.4pt}
\renewcommand{\arraystretch}{1.08}

\begin{tabular}{c l cccccc|cc}
\toprule
\multirow{2}{*}{\textbf{Models}}
&
\multicolumn{1}{c}{\multirow{2}{*}{\textbf{Methods}}}
&
\multicolumn{6}{c|}{\textbf{Accuracy (\%)}}
&
\multicolumn{2}{c}{\textbf{Overall}}
\\
\cmidrule(lr){3-8}
\cmidrule(l){9-10}
&
&
\textbf{MMLU}
&
\textbf{GPQA-D}
&
\textbf{GSM8K}
&
\textbf{MATH-500}
&
\textbf{MedQA}
&
\textbf{LogiQA2.0}
&
\textbf{Avg.\ Acc.}
&
\textbf{Avg.\ Tokens}
\\
\midrule

\multirow{3}{*}{%
  \rotatebox[origin=c]{90}{\footnotesize GPT}%
}
&
CoT~\cite{wei2022chain}
& 87.04
& 64.65
& 95.60
& 87.17
& 90.02
& 76.08
& 83.43
& 738
\\
&
Self-Refine~\cite{madaan2023self}
& 87.11
& 67.68
& 94.47
& 88.60
& 90.97
& 76.40
& 84.21
& 3{,}501
\\
&
Self-Consistency~\cite{wangself}
& 87.92
& 68.18
& 95.83
& 91.00
& 90.57
& 76.59
& 85.02
& 3{,}529
\\

\midrule

\multirow{3}{*}{%
  \rotatebox[origin=c]{90}{\footnotesize DS}%
}
&
CoT~\cite{wei2022chain}
& 86.50
& 61.62
& 96.13
& 88.55
& 85.39
& 74.75
& 82.16
& 984
\\
&
Self-Refine~\cite{madaan2023self}
& 86.70
& 60.61
& 95.52
& 86.00
& 88.14
& 75.83
& 82.13
& 3{,}284
\\
&
Self-Consistency~\cite{wangself}
& \underline{88.81}
& 61.11
& \textbf{96.89}
& 92.00
& 85.62
& 79.33
& 83.96
& 3{,}466
\\

\midrule

\multirow{6}{*}{%
  \rotatebox[origin=c]{90}{\footnotesize GPT+DS}%
}
&
MAD~\cite{du2024improving}
& 88.09
& 65.61
& 96.59
& 91.60
& 91.12
& 79.70
& 85.45
& 25{,}961
\\
&
HCP-MAD~\cite{liu2026hcp}
& 88.37
& 66.16
& 96.66
& \underline{92.20}
& \underline{91.20}
& 79.77
& \underline{85.73}
& 4{,}938
\\
&
DAAO~\cite{su2026daao}
& 88.65
& 65.66
& \underline{96.74}
& 86.00
& 90.57
& 77.85
& 84.25
& 8{,}856
\\
&
BAMAS~\cite{yang2026bamas}
& 86.70
& 61.42
& 95.38
& 87.40
& \underline{91.20}
& 80.09
& 83.70
& 1{,}217
\\
&
AnyMAC~\cite{wang2025anymac}
& 86.91
& \underline{68.69}
& 95.60
& 90.60
& 86.72
& \textbf{80.47}
& 84.83
& 8{,}964
\\
\rowcolor{gray!10}
&
\textbf{TreeCredit(Ours)}
& \textbf{89.88}
& \textbf{69.19}
& 96.51
& \textbf{92.80}
& \textbf{91.44}
& \underline{80.15}
& \textbf{86.66}
& 2{,}193
\\

\bottomrule
\end{tabular}

\caption{Comparison across six reasoning benchmarks.
For each accuracy column, the best result is shown in {bold} and the second-best is {underlined}.
GPT and DS denote GPT-4.1-mini and DeepSeek-V3-0324, respectively; all collaboration and routing methods use the same GPT+DS heterogeneous pair.
Avg.\ Acc. is the macro-average over the six benchmarks, and Avg.\ Tokens is the average token consumption per query.}
\label{tab:treecredit_main}
\end{table*}

\section{Experiments}

\subsection{Experimental Setup}


\textbf{Datasets:}
We evaluate TreeCredit on six benchmarks:
MMLU~\cite{hendrycksmeasuring} and GPQA-Diamond~\cite{rein2024gpqa} for general and scientific knowledge,
GSM8K~\cite{cobbe2021training} and MATH-500~\cite{hendrycks2021measuring} for mathematical reasoning,
MedQA~\cite{jin2020medqa} for medical question answering, and LogiQA2.0 for logical reasoning.
The standard evaluation split and answer-extraction protocol for each benchmark.

\noindent\textbf{Models and Collaboration Setting.}
Our core experiments use GPT-4.1-mini and DeepSeek-V3-0324 as a heterogeneous agent pair.
Single-agent baselines evaluate each model independently, whereas all collaboration and routing methods use the same pair.

\noindent\textbf{Baselines.}
For single-agent reasoning, we include Chain-of-Thought (CoT)~\cite{wei2022chain},
Self-Refine~\cite{madaan2023self}, and five-sample Self-Consistency~\cite{wangself} for both backbones.
For collaboration, we compare with fixed Multi-Agent Debate (MAD)~\cite{du2024improving} and the rule-based progressive HCP-MAD workflow.
We further include DAAO~\cite{su2026daao} as a query-level router and BAMAS~\cite{yang2026bamas} and AnyMAC~\cite{wang2025anymac} as adaptive-routing baselines.

\noindent\textbf{Evaluation Metrics.}
We report task accuracy and token consumption per query.
Avg.\ Acc. is the unweighted mean of the six benchmark accuracies, while Avg.\ Tokens is the unweighted mean of the token averages.

\noindent\textbf{Implementation Details.}
For each benchmark, we randomly sample 100 labeled questions from the training split for \emph{Shared-Prefix Tree Expansion}. 
For each benchmark, we train a separate TreeCredit router using 100 labeled examples sampled exclusively from the official training set. 
All reported results are evaluated on a disjoint test set. 
For each question, we construct (K=2) independent collaboration trees using different decoding seeds, yielding 200 trees per benchmark. 
Note that the trees are not aggregated.
Each collaboration trajectory contains at most ($R_{\max}$=5) debate rounds. 
The router is implemented as a two-layer MLP with a hidden dimension of 128. Its 398-dimensional input concatenates a 384-dimensional query embedding produced by the frozen \texttt{all-MiniLM-L6-v2} encoder with a 14-dimensional compact collaboration-state representation, and its output layer produces logits for the three operators: \emph{Accept}, \emph{Debate}, and \emph{Vote}.


\subsection{Main Results}
We compare TreeCredit with representative single-agent, multi-agent, and learned routing methods across six reasoning benchmarks, with the main results summarized in Table~\ref{tab:treecredit_main}.
Among single-agent baselines, Chain-of-Thought requires the fewest tokens but achieves relatively lower accuracy. On GPT-4.1-mini, Self-Consistency increases the average accuracy from 83.43\% to 85.02\%, at the cost of raising token consumption from 738 to 3,529. 
In comparison, TreeCredit achieves a higher average accuracy of 86.66 \% while reducing token consumption to 2,193. 
These results suggest that state-adaptive collaboration provides a more favorable accuracy-cost trade-off than repeatedly sampling independent solutions from a single agent.



TreeCredit also compares favorably with conventional multi-agent reasoning methods. Fixed Multi-Agent Debate achieves an average accuracy of 85.45\% with 25{,}961 tokens, while HCP-MAD substantially reduces the cost to 4{,}938 tokens and obtains 85.73\% accuracy. Relative to HCP-MAD, TreeCredit further improves average accuracy by 0.93 percentage points while reducing token consumption by 55.59\%. These results indicate that state-conditioned routing preserves the reasoning benefits of multi-agent collaboration while avoiding unnecessary debate and aggregation.

Among learned routing approaches, DAAO and AnyMAC obtain average accuracies of 84.25\% and 84.83\% with 8{,}856 and 8{,}964 tokens, respectively, whereas TreeCredit reaches 86.66 \% with only 2{,}193 tokens. BAMAS achieves the lowest inference cost at 1{,}217 tokens, but its average accuracy is 83.70\%, 2.96 percentage points below TreeCredit. Its low cost is associated with a conservative collaboration policy that uses approximately one to two model calls per query, which limits additional reasoning on difficult instances. 


\begin{figure}
\centering
\includegraphics[
    width=0.9\columnwidth,
    height=0.25\textheight
]{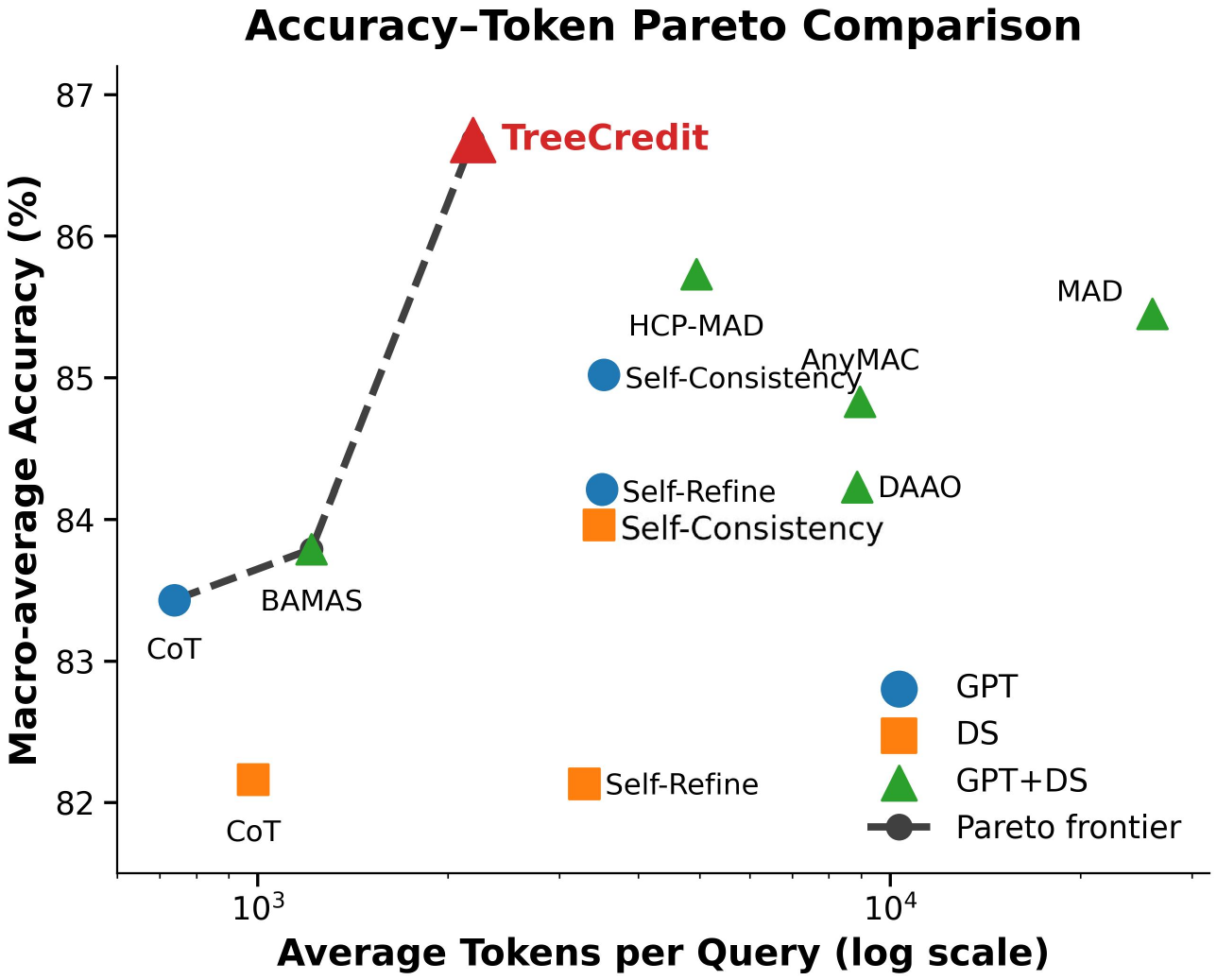}
\caption{Accuracy-Cost comparison using the average results over six benchmarks.}
\label{fig:pareto_accuracy_tokens}
\end{figure}


Figure~\ref{fig:pareto_accuracy_tokens} compares average accuracy and token consumption, with better operating points located toward the upper-left. TreeCredit lies on the empirical Pareto frontier and achieves the highest average accuracy among methods with complete results. It strictly dominates Multi-Agent Debate, HCP-MAD, DAAO, AnyMAC, and Self-Consistency, which are both less accurate and more costly. Chain-of-Thought and BAMAS use fewer tokens but incur substantial accuracy losses; for example, BAMAS saves 976 tokens but trails TreeCredit by 2.96 percentage points. Overall, TreeCredit achieves the most favorable accuracy--cost trade-off by allocating additional collaboration only when its expected benefit justifies the cost.

\begin{figure}
\centering
\includegraphics[width=0.98\columnwidth]{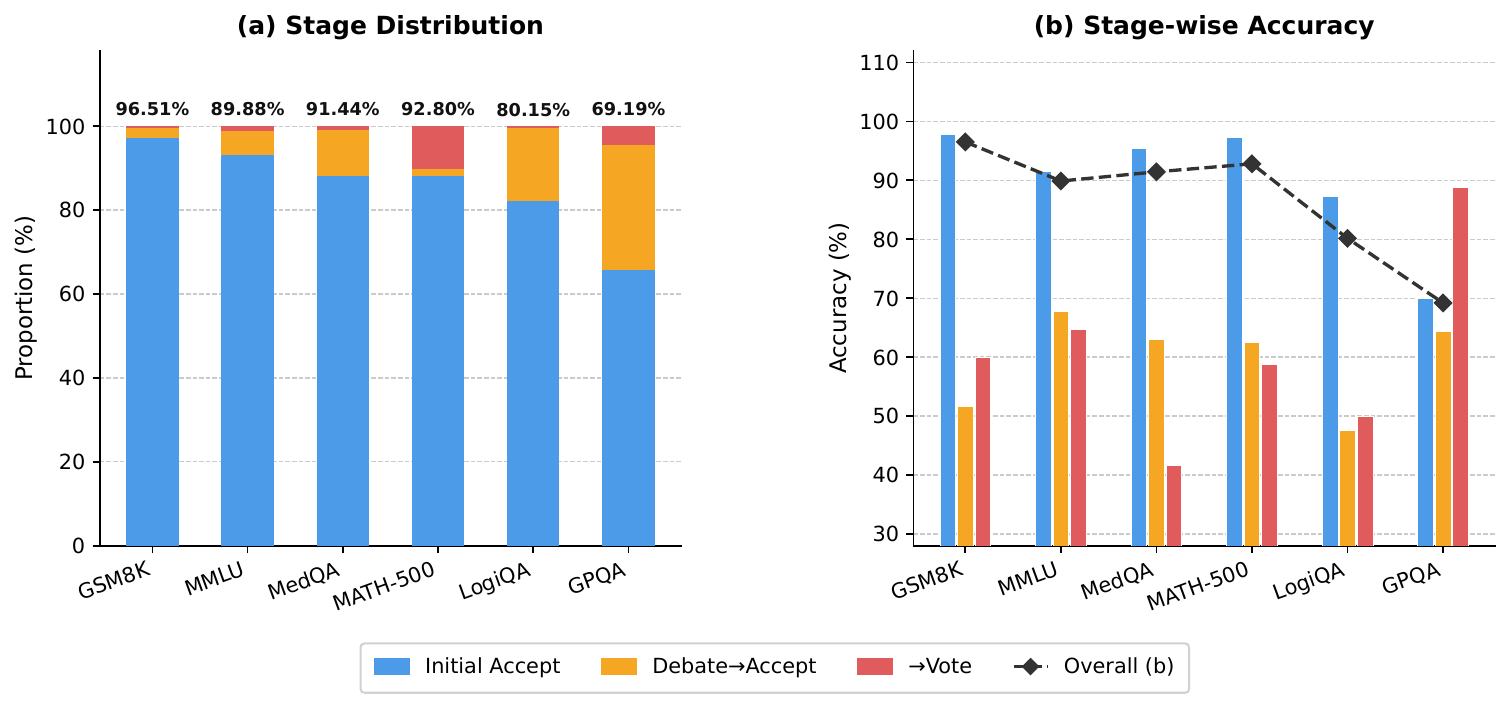}
\caption{Stage-wise routing behavior of TreeCredit. The left panel shows the proportion of queries terminated by initial Accept, Debate$\rightarrow$Accept, or Vote. The right panel reports the conditional accuracy of each routed subset, with the dashed curve indicating overall accuracy.}
\label{fig:stage_distribution}
\end{figure}




\subsection{Routing and Outcome Analysis}

\paragraph{Progressive Routing Behavior.} 
Figure~\ref{fig:stage_distribution} shows that TreeCredit terminates most queries via initial accept at round 0, including 97.27\% of GSM8K, 93.21\% of MMLU, 88.22\% of MedQA,
and 88.20\% of MATH-500 instances. 
These early-accepted subsets are highly reliable, with conditional accuracies of
97.74\%, 91.52\%, 95.46\%, and 97.28\%, respectively.
More difficult benchmarks require broader collaboration:
GPQA-D routes 29.79\% of queries through Debate and 4.55\% to Vote, while LogiQA2.0 routes 17.49\% through Debate; 
MATH-500 invokes Vote for 10.20\% of its queries. 
Because later decisions apply only to instances unresolved at initial round,
their conditional accuracies characterize harder router-selected subsets rather than the intrinsic quality of Debate or Vote.
Overall, TreeCredit achieves 69.19\%--96.51\% accuracy while terminating reliable cases
early and reserving additional collaboration for harder queries.

\begin{table}[t]
\centering
\footnotesize
\setlength{\tabcolsep}{2pt}
\begin{tabular}{lcccc}
\toprule
Dataset
& $\times\!\rightarrow\!\checkmark$ (\%)
& $\checkmark\!\rightarrow\!\times$ (\%)
& $\checkmark\!\rightarrow\!\checkmark$ (\%)
& $\times\!\rightarrow\!\times$ (\%) \\
\midrule
GPQA-D    & 8.08 & 1.01 & 61.11 & 29.80 \\
LogiQA2.0 & 6.11 & 1.46 & 74.05 & 18.38 \\
MedQA     & 5.03 & 2.36 & 86.41 &  6.21 \\
MMLU      & 4.38 & 1.18 & 85.50 &  8.94 \\
MATH-500  & 3.60 & 1.00 & 89.20 &  6.20 \\
GSM8K     & 1.44 & 0.30 & 95.07 &  3.18 \\
\bottomrule
\end{tabular}
\caption{Correctness transitions from the initial reference
prediction to the final TreeCredit prediction over all queries.
Datasets are ordered by the
$\times\!\rightarrow\!\checkmark$ rate in descending order.}
\label{tab:treecredit_transition_compact}
\end{table}

\paragraph{Analysis of Correctness Transitions.}
Table~\ref{tab:treecredit_transition_compact} compares the
initial reference prediction with the final TreeCredit
prediction for every query. When the two heterogeneous agents
agree at initial round, their shared answer is used as the initial
reference; otherwise, Agent A's answer is used. Across the six
datasets, TreeCredit corrects 1.44\%--8.08\% of all queries
that are initially incorrect, while changing only
0.30\%--2.36\% of all queries from correct to incorrect.
The correction rate is higher on challenging datasets such as
GPQA-D and LogiQA2.0, whereas smaller rates on high-accuracy
datasets mainly reflect fewer initial errors. Since
$\times\!\rightarrow\!\checkmark$ exceeds
$\checkmark\!\rightarrow\!\times$ on every benchmark,
TreeCredit yields a positive net accuracy gain across all six
datasets.

\subsection{Ablation Study}
\label{sec:ablation}

\begin{figure}
    \centering
    \includegraphics[width=0.8\columnwidth]
    {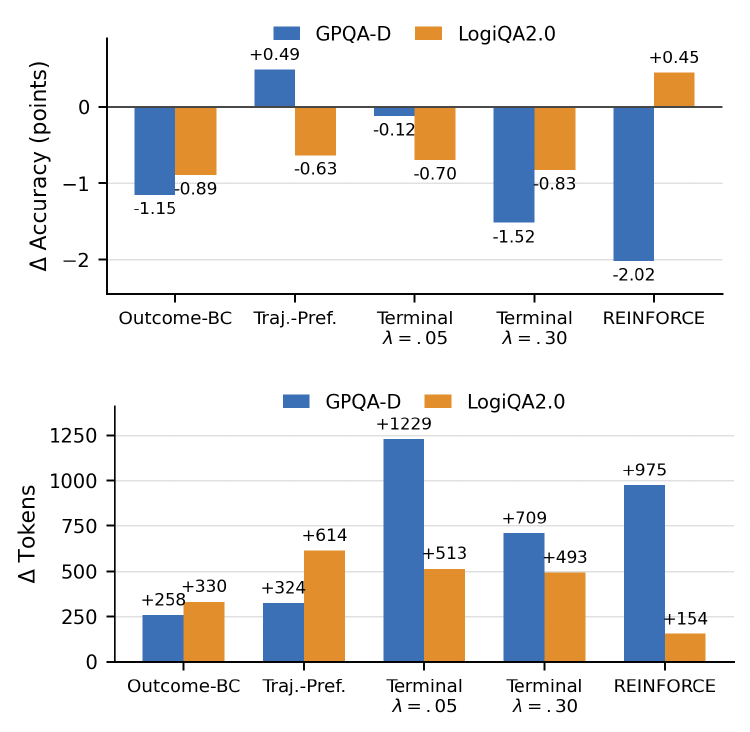}
    \caption{
    Supervision ablations relative to TreeCredit on GPQA-D and
    LogiQA2.0. Each bar reports \text{Variant} vs. \text{TreeCredit}.
    Positive $\Delta$Accuracy indicates higher accuracy, whereas
    positive $\Delta$Tokens indicates higher inference cost.
    }
    \label{fig:supervision_ablation}
\end{figure}

We separate the ablation study into supervision and design analyses because they examine complementary aspects of TreeCredit. Figure~\ref{fig:supervision_ablation} compares alternative supervision strategies with the full method under the same model pair, operator space, and offline data budget. 
The only difference is how router supervision is constructed, allowing us to isolate whether shared-prefix comparisons provide more informative credit than complete-trajectory targets or delayed terminal rewards.
Otherwise, Table~\ref{tab:design_ablation} instead retains shared-prefix supervision and removes one design component at a time. 
This controlled setting evaluates the independent contributions of correctness-prioritized credit assignment and state-conditioned routing.


\paragraph{Outcome-Supervised Trajectory.}
This variant assigns positive or negative labels to all actions according to the final trajectory outcome, without comparing operators from the same state. 
Compared with TreeCredit, it lowers GPQA-D and LogiQA2.0 accuracy by 1.15 and 0.89
points while increasing token usage by 258 and 330.
Because successful trajectories may contain redundant actions and failed ones also may include useful intermediate decisions, terminal labels provide unreliable local credit. 
TreeCredit instead compares sibling operators under a shared prefix to obtain state-matched supervision.

\paragraph{Trajectory-Level Preference.}
Trajectory-Level Preference ranks complete trajectories by final correctness and total cost, but still compares entire action sequences rather than operators at a shared state. 
Relative to TreeCredit, it improves GPQA-D accuracy by 0.49 points while using 324 more tokens; on LogiQA2.0, it lowers accuracy by  0.63 points and adds 614 tokens. 
Because trajectories may differ at multiple routing steps, their overall ranking cannot identify which decision caused the gain or loss. 
It may also favor a long correct trajectory even when earlier acceptance would achieve the same correctness at lower cost. 
TreeCredit avoids this confounding by comparing candidate operators after an identical shared prefix.


\paragraph{Terminal-Reward Preference.}
Terminal-Reward Preference ranks complete trajectories using a scalar reward that combines final correctness and token cost with coefficient $\lambda$, testing whether scalarization can replace TreeCredit's correctness-prioritized suffix credit. 
With $\lambda=0.05$, accuracy on GPQA-D and LogiQA2.0 decreases by 0.12 and 0.70 points, while token usage increases by 1,229 and 513. 
Raising $\lambda$ to 0.30 saves 520 and 20 tokens, yet causes additional accuracy drops of
1.40 and 0.13 points. 
It therefore remains both less accurate and more expensive than TreeCredit. 
These results show that scalar rewards are sensitive to $\lambda$: 
small values under-penalize cost, while larger values sacrifice correctness before matching TreeCredit's efficiency.
Moreover, tuning $\lambda$ changes only the global accuracy--cost trade-off
and does not resolve intermediate credit assignment.


\paragraph{REINFORCE Router.}
REINFORCE uses the same router, state representation, and depth constraint as TreeCredit, but learns from delayed trajectory-level rewards.
Compared with TreeCredit, it lowers GPQA-D accuracy by 2.02 points while using 975 more tokens; on LogiQA2.0, it gains 0.45 points but requires 154 additional tokens, yielding no consistent accuracy--cost advantage.
Because all actions in a trajectory share the same return, unnecessary later actions may be reinforced, while useful intermediate decisions can be penalized by subsequent errors. TreeCredit instead compares sibling operators at the same state, providing more direct supervision for choosing \emph{Accept}, \emph{Debate}, or \emph{Vote}.


\begin{table}[t]
\centering
\footnotesize
\setlength{\tabcolsep}{3.0pt}
\renewcommand{\arraystretch}{1.05}
\resizebox{\columnwidth}{!}{
\begin{tabular}{lcccc}
\toprule
\multirow{2}{*}{\textbf{Variant}}
& \multicolumn{2}{c}{\textbf{GPQA-D}}
& \multicolumn{2}{c}{\textbf{LogiQA2.0}} \\
\cmidrule(lr){2-3}\cmidrule(lr){4-5}
& \textbf{Acc.} & \textbf{Tok.}
& \textbf{Acc.} & \textbf{Tok.} \\
\midrule
\rowcolor{gray!10}
\textbf{TreeCredit}
& \textbf{69.19} & \textbf{4{,}107}
& \textbf{80.15} & \textbf{1{,}812} \\
w/o Correctness-Prioritized Credit
& 64.95 & 4{,}573
& 78.63 & 2{,}049 \\
w/o State-Conditioned Routing
& 68.04 & 5{,}368
& 79.07 & 2{,}166 \\
\bottomrule
\end{tabular}
}
\caption{Ablation of TreeCredit design components.}
\label{tab:design_ablation}
\end{table}

\paragraph{Design Components.}
Table~\ref{tab:design_ablation} further isolates the contributions of two key components within the shared-prefix framework. 
Removing correctness-prioritized credit produces the largest performance drop: accuracy on GPQA-D and LogiQA2.0 decreases by 4.24 and 1.52 points, respectively, while token consumption increases by 466 and 237. 
Without the correctness-first ordering, lower-cost actions may be preferred even when they reduce the likelihood of reaching a correct answer. 
Removing state-conditioned routing also lowers accuracy by 1.15 and 1.08 points and increases token usage by 1{,}261 and 354. 
Because this variant cannot adapt to evolving agreement, confidence, and interaction history, it often continues reasoning in states that could already be terminated. 
Together, these results show that correctness-prioritized credit preserves answer quality, while state-conditioned routing enables efficient, instance-specific allocation of collaboration.

\section{Conclusion}
We propose TreeCredit, a shared-prefix credit assignment framework that derives correctness-prioritized, state-local operator preferences to train a lightweight router for adaptive multi-agent reasoning. Across six benchmarks, TreeCredit achieves a favorable accuracy–cost trade-off by improving reasoning performance while substantially reducing token usage. Current limitations include its restricted Accept–Debate–Vote operator set and reliance on benchmark-specific labeled data; future work will explore richer operators, cross-domain routing, and uncertainty-aware credit estimation.





\bibliography{main}


\end{document}